\documentclass{article}
\usepackage{spconf,amsmath,graphicx,hyperref}
\usepackage{booktabs,multirow,makecell}
\usepackage{microtype}
\usepackage[table]{xcolor}
\usepackage{cite}
\definecolor{revisionblue}{RGB}{0,87,184}
\hypersetup{hidelinks}

\title{Holistic Parallel Supervision for Expressive Speech to Speech Translation}
\name{\shortstack{%
Hao Zhang, Ismail Rasim Ulgen, Bismarck Bamfo Odoom,\\
Rui Liu, Zihan Zhang, Berrak Sisman, Philipp Koehn}}
\address{Center for Language and Speech Processing, Johns Hopkins University, MD, USA}

\begin{document}

\maketitle
\begingroup
\renewcommand{\thefootnote}{}
\footnotetext{\raggedright\textbf{Codes and demo:}~{https://hzted.github.io/truly-e2e-expressive-s2st-demo} \\
Due to licensing constraints, the dataset itself cannot be shared.}
\endgroup
\begin{abstract}

Expressiveness is central to speech-to-speech translation (S2ST), yet existing datasets rarely align source and target speech jointly in content, prosody, timing, and speaker identity. Consequently, current systems often rely on indirect conditioning or transfer mechanisms. To obtain this supervision, we introduce a training-data construction strategy that combines professional dubbing, which provides human-performed correspondence in content, prosody, and timing, with multilingual voice conversion to reduce speaker mismatch. We evaluate this supervision using existing expressive S2ST systems and DirectS2ST, an end-to-end method developed in this work that predicts the complete target-speech representation directly from source speech. Experiments on English--German and English--Spanish show that holistic parallel supervision improves expressive realization across all evaluated systems. DirectS2ST exhibits the largest gains in content, prosody, timing, speaker similarity, and human-rated expressiveness. To our knowledge, this work provides the first systematic investigation of holistic parallel supervision for expressive S2ST.

\end{abstract}


\section{Introduction}
\label{sec:intro}

Speech-to-speech translation (S2ST) should preserve not only what is said but also how it is said \cite{tsiamas2024speech}. Prosodic cues such as stress and intonation convey meaning and intent \cite{cutler1997prosody}, while vocal style and speaker identity contribute to expressive communication \cite{jia2022translatotron2}. However, existing methods and resources largely emphasize lexical accuracy, which alone cannot preserve these properties. Learning expressive translation requires parallel speech pairs aligned in content, prosody, timing, and speaker identity; semantically correct but expressively mismatched targets do not demonstrate how the source expression should be realized in another language.

Such holistic parallel data are scarce because collecting cross-lingual recordings from the same speaker with corresponding expressive delivery is costly. Existing resources therefore leave at least one dimension unmatched. Synthetic speech can reproduce source-speaker identity but may lack expressive delivery \cite{jia2022cvss}; mined data provide semantic parallelism at scale but often have very limited speaker and expressive correspondence \cite{barrault2023seamless}; and professional dubbing provides strong semantic, prosodic, and temporal correspondence but uses different speakers \cite{xie2026prosodylosttranslationfinegrained}. Thus, existing S2ST systems rarely receive supervision aligned across all four dimensions. As a result, expressive S2ST methods often rely on indirect strategies, including replacing target-derived training conditions with source-derived conditions at inference or transferring speaker and prosodic information through separate mechanisms \cite{le2024transvip,ulgen2026direct,choi2025dubs2st,cheng2026uniss}. Although effective, these approaches do not learn content and expressive realization jointly through direct paired-speech supervision.

\begin{figure*}[!t]
    \centering
    \resizebox{\textwidth}{!}{%
        \begin{tabular}{@{}c@{\hspace{1pt}}c@{}}
            \includegraphics[height=2.30in]{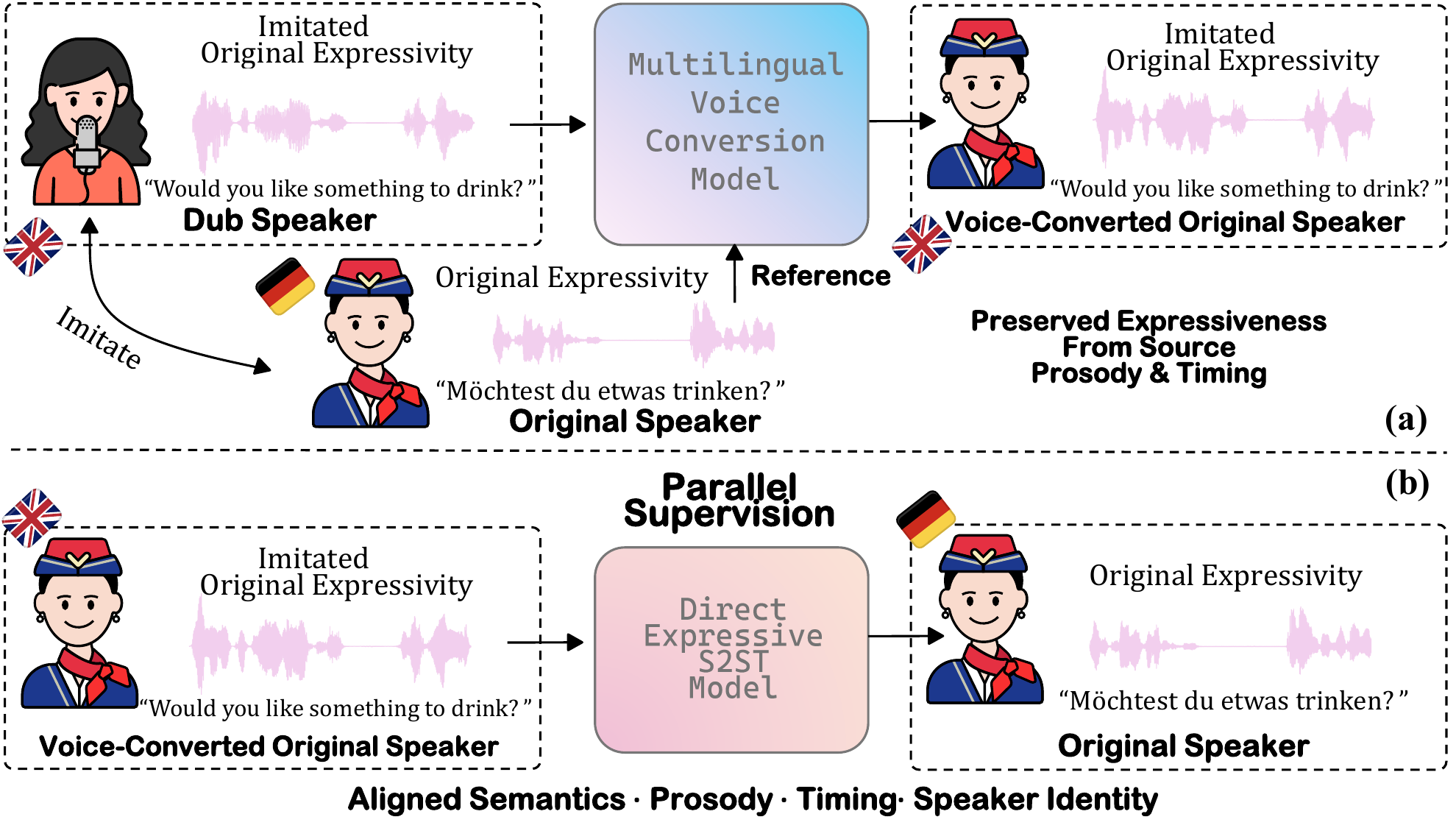} &
            \includegraphics[height=2.30in]{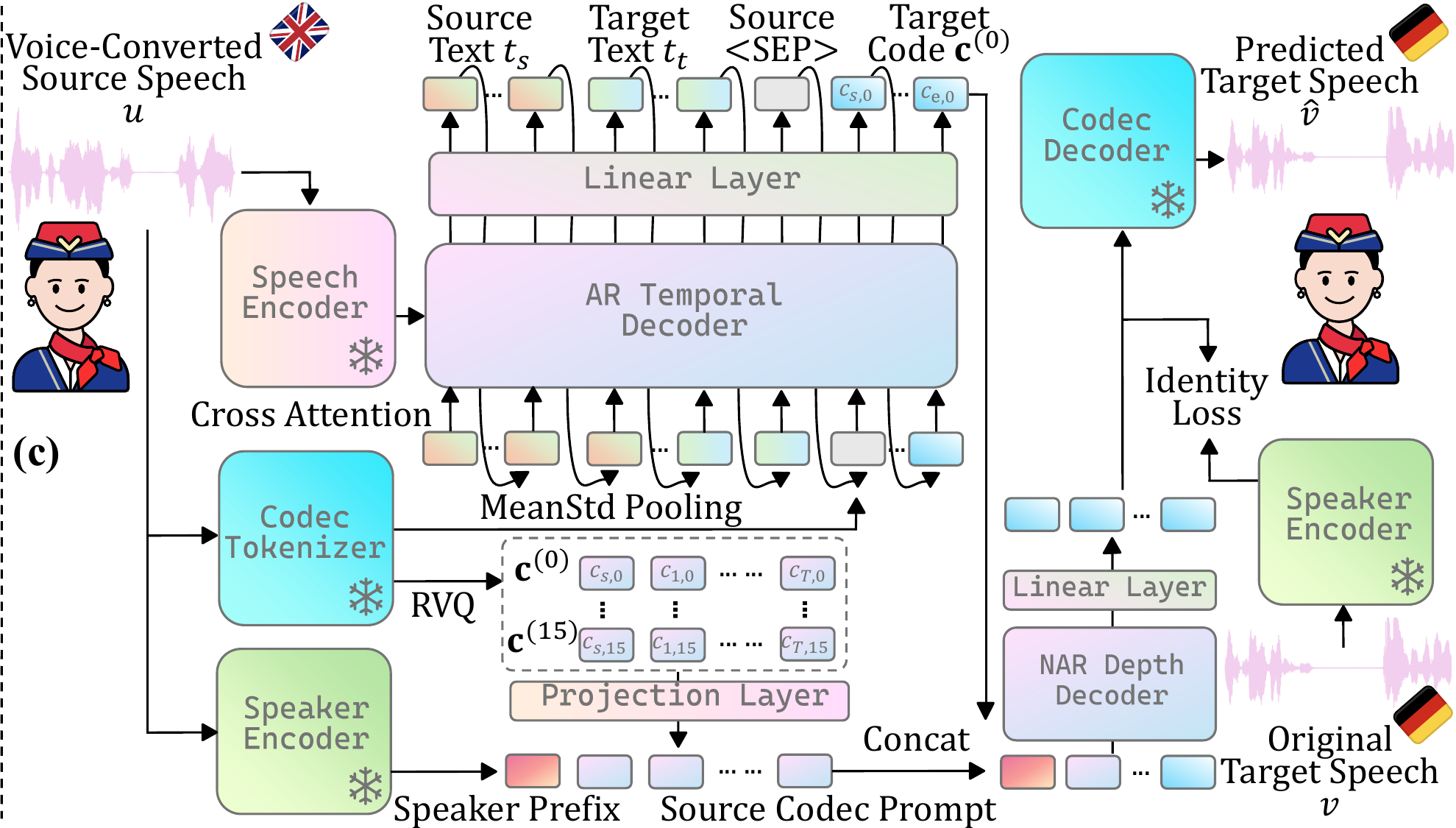} \\[-1mm]
                   \end{tabular}%
    }
    \vspace{-4mm}
\caption{Overview of the holistic parallel supervision:
(a) \textsc{VC-Dub} Construction; (b) Parallel Supervision; (c) DirectS2ST.}
    \label{fig:framework}
    \vspace{-5mm}
\end{figure*}

In this work, we investigate holistic parallel supervision for expressive S2ST. Because collecting such data from scratch is costly, we combine professional dubbing with multilingual voice conversion (VC). Dubbing provides human-performed correspondence in content, prosody, and timing, while VC reduces speaker mismatch. We evaluate this supervision using existing expressive S2ST methods and DirectS2ST, an end-to-end model developed in this work that predicts the complete target-speech representation directly from source speech. DirectS2ST tests whether direct modeling can exploit holistic supervision more fully. Our contributions are: (i) the first proof-of-concept study, to our knowledge, of holistic parallel supervision for expressive S2ST; (ii) a training-data construction strategy combining professional dubbing and multilingual VC, evaluated using existing methods and DirectS2ST; and (iii) consistent improvements in expressive preservation on En--De and En--Es, supported by human evaluation.

\section{holistic Parallel Supervision For S2ST}
\label{sec:method}

We instantiate our holistic parallel supervision hypothesis with two components:
(1) We construct holistic parallel cross-lingual speech pairs matched in
content, prosody, timing, and speaker identity for training. (2) We study holistic parallel supervision in two settings:
existing expressive S2ST systems, and DirectS2ST, which examines whether holistic paired
supervision can be exploited more directly through complete
target-speech prediction, rather than through target-derived
training conditions or separate expressive-transfer mechanisms.
\vspace{-3mm}
\subsection{Training Dataset Construction Method: \textsc{VC-Dub}}
\label{sec:vcdub}

\noindent\textbf{Human-performed expressive correspondence.}
Professional dubbing renders the original dialogue in another
language while seeking to convey its expressive intent within
the same audiovisual context.
Beyond translated content, dubbed performances can reflect
the original emphasis, intonation, and
temporal organization \cite{brannon-etal-2023-dubbing}.
We therefore use professional dubbing from entertainment television episodes
spanning multiple genres (e.g., drama, comedy, action) as human-performed cross-lingual
supervision rather than requiring identical
pitch contours or frame-level timing.
The remaining limitation is speaker mismatch, since the two
performances are generally produced by different actors.

\noindent\textbf{Pair preparation.}
We construct English-to-German and English-to-Spanish pairs
from German or Spanish original tracks and their English dubs. We apply multiple preprocessing steps to make raw professional dubbing data suitable for speech translation training, in terms of semantic accuracy, speech quality and removal of background music/noise. 
We use Speech Vecalign \cite{meng-koehn-2025-speech}
to semantically align speech segments between the
corresponding original and dubbed audio tracks.
Before VC, we process both sides with
ClearerVoice-Studio for speech enhancement and Demucs for
vocal separation to have single speaker utterances \cite{zhao2025clearervoice,rouard2023hybrid}.
We then check the expected languages using MMS-LID,
reject multi-speaker segments using Sortformer
\cite{park2024sortformer}, and filter low-quality pairs
using the speech-quality
predictor DNSMOSPro \cite{cumlin2024dnsmos}. All scripts, thresholds, and implementation details are released. 

\noindent\textbf{Speaker matching and target supervision.}
Let $x^{a,o}$ denote a cleaned original utterance in language
$a$ spoken by speaker $o$, and $y^{b,d}$ its cleaned dubbed
counterpart in language $b$ spoken by actor $d$.
We apply multilingual zero-shot voice conversion, instantiated
with Seed-VC \cite{liu2024zero}, to the dubbed side:
\vspace{-1.1mm}
\begin{equation}
\tilde{y}^{b,o}
=
\mathcal{V}_{\mathrm{multi}}
\left(y^{b,d};x^{a,o}\right).
\label{eq:vc}
\vspace{-1.1mm}
\end{equation}
Here, $y^{b,d}$ provides translated content and performed
delivery, while $x^{a,o}$ provides the speaker reference.
Thus, VC reduces the speaker mismatch. Since the VC method is designed to preserve the source linguistic content and prosody \cite{liu2024zero}, the semantic and expressive parallelism is largely preserved. 


We form the S2ST training pair in the reverse direction
of the original dubbing process:
\vspace{-1.3mm}
\begin{equation}
(u,v)
=
\left(\tilde{y}^{b,o},x^{a,o}\right)
\in
\mathcal{D}_{\mathrm{\textsc{VC-Dub}}}^{b\rightarrow a}.
\label{eq:vcdub}
\vspace{-1.2mm}
\end{equation}
The VC applied English dub becomes the source, while the
original German or Spanish recording remains the target during supervision.
Crucially, the target retains the human performance without TTS or VC resynthesis, so content,
prosody, timing, and speaker identity are supervised jointly
to predict real speech distribution as target.
Applying VC only to the source avoids introducing VC-generated artifacts into target speech supervision. Pair-level analysis verifies the intended correspondence before
we evaluate its effect on downstream S2ST (see Section \ref{sec:results}).
\vspace{-3mm}
\subsection{DirectS2ST}
\label{sec:model}

DirectS2ST uses a Moshi-inspired temporal--depth architecture
\cite{defossez2024moshi} to test whether holistic parallel
supervision can be exploited more fully through direct
source-to-target expressive learning.
Given source speech $u$, it predicts all 16 Mimi codec streams
of the paired human recording $v$. A frozen w2v-BERT~2.0 encoder \cite{barrault2023seamless}
provides continuous source memory through cross-attention.
The autoregressive (AR) temporal decoder generates source text,
target text, and the first-level target codec stream
$\mathbf{c}^{(0)}$ with a shared output head.
Inspired by TransVIP \cite{le2024transvip}, the
\texttt{<SEP>} embedding at the text-to-codec boundary is replaced
with an acoustic prompt obtained by mean--standard-deviation
pooling of continuous Mimi features.
Unlike TransVIP's target-derived training prompt, ours is
extracted from the source speech.

The remaining streams are predicted by a shared
non-autoregressive (NAR) depth Transformer.
Each level conditions on lower target codebooks and a source
codec prefix, with a projected embedding from a frozen CAMPPlus
speaker encoder \cite{wang23ha_interspeech} prepended as the
speaker-identity condition. We further apply a speaker-identity loss by
projecting mean-pooled NAR hidden states to the CAMPPlus space
and matching them by cosine loss to the paired target-speech
embedding, reinforcing speaker learning in the depth decoder.

Let $\mathbf{z}$ denote the AR sequence containing source text $t_{s}$,
target text $t_{t}$, and $\mathbf{c}^{(0)}$.
Prediction is factorized as
\setlength{\abovedisplayskip}{4pt}
\setlength{\belowdisplayskip}{3pt}
\begin{equation}
\begin{aligned}
p_{\mathrm{AR}}(\mathbf{z}\mid u)
&=
\prod_i p(z_i\mid\mathbf{z}_{<i},u),\\
p_{\mathrm{NAR}}(\mathbf{c}^{(1:15)}
\mid u,\mathbf{c}^{(0)})
&=
\prod_{q=1}^{15}\prod_{t=1}^{T}
p(c_{t,q}\mid\mathbf{c}^{(<q)},\mathbf{P}_u),
\end{aligned}
\label{eq:directs2st}
\end{equation}

where $u$ is source speech and $\mathbf{P}_u$ denotes the source codec and speaker prefix. The NAR stage predicts target frames in parallel within each codebook while proceeding sequentially across codebook levels. The predicted source and target text, $t_s,t_t$ serve the purpose of helping semantic grounding and guidance for translation while source speech condition $u$ guides all the aspects including semantic, speaker identity, and prosody. 
Using source speech as the only and direct condition while predicting all these aspects realizes end-to-end and holistic parallel supervision during training. \vspace{-1mm}

\section{Experimental Setup}

\label{sec:exp}

\noindent\textbf{Training Data.} We compare \textsc{VC-Dub} with CVSS-T \cite{jia2022cvss}, a widely used S2ST corpus with synthetic voice-transferred targets, using approximately matched training-set sizes (Table \ref{tab:data_statistics}). For Dub-S2ST, we also evaluate its proposed speed-adapted CVSS-T variant (CVSS-T$^{\mathrm{S}}$) \cite{choi2025dubs2st}, which improves source--target speaking-rate correspondence but not other expressive dimensions. To isolate the effect of speaker matching, we train DirectS2ST on the same dubbed pairs without VC (\textsc{Raw-Dub}). Source and target transcripts are generated using Whisper-large-v3 \cite{radford2023robust} and used only as auxiliary training targets, not as inference-time inputs.

\noindent\textbf{Evaluation Data.}
We evaluate on mExpresso \cite{nguyen2023expresso, barrault2023seamless}, an expressive multilingual dataset independent of our training corpus. Each of the En--De and En--Es evaluations contains 504 examples evenly distributed across seven expressive conditions. Dub-S2ST is evaluated only on En--Es because its mHuBERT unit representation does not support German.

\noindent\textbf{Metrics.}
We evaluate content, prosody, timing, speaker identity, and naturalness. ASR-BLEU uses Whisper-large-v3 transcripts and SacreBLEU \cite{radford2023robust,post-2018-call}, while BLASER2.0-QE measures source--output semantic consistency \cite{dale-costa-jussa-2024-blaser}. AutoPCP (A.PCP) measures utterance-level prosody similarity, and STOPES \cite{barrault2023seamless} measures speaking-rate correlation (Rate), pause similarity (Pause), and speaker similarity (Vsim) between source and translated speech. NISQA-TTS (NAT) estimates naturalness \cite{mittag2020syntheticnaturalness}. For VC diagnostics, WER and CER between transcripts before and after VC measure content preservation.

\noindent\textbf{Human evaluation.} Nine listeners proficient in English and Spanish rate the expressiveness of outputs from six system configurations on a 5-point expressiveness MOS scale, considering emphasis, rhythm, timing, and intonation. The test uses 10 En--Es source utterances sampled across seven expressive conditions, with the source provided as reference.

\noindent\textbf{S2ST Methods.} In addition to DirectS2ST, we evaluate holistic parallel supervision with TransVIP \cite{le2024transvip} and Dub-S2ST \cite{choi2025dubs2st}. We train the official TransVIP implementation on our data using its default settings. For Dub-S2ST, whose original speech-to-unit module showed limited generalization in preliminary experiments, we replace that module with the S2U model of \cite{popuri22_interspeech}. We train both the S2U and U2S components on our data, with
the U2S module initialized from Dub-S2ST's pretrained
CosyVoice-based checkpoint rather than from scratch.
 
\noindent\textbf{Training \& Architecture Details.}
The temporal and depth Transformers have 12 and 8 layers, respectively, with a shared configuration of 1024 hidden dimensions, 16 attention heads, 4096 feed-forward dimensions, and 0.1 dropout, yielding a 1.2B-parameter DirectS2ST model with randomly initialized decoders. We use a direction-specific 8k SentencePiece vocabulary \cite{kudo-richardson-2018-sentencepiece}, the first 16 codebooks of the frozen pretrained Mimi codec (2,048 entries each), and a frozen CAMPPlus speaker encoder \cite{wang23ha_interspeech}. Separate En--De and En--Es models are trained on one 141-GB NVIDIA H200 GPU with a batch size of 8 for approximately 2--3 days each. Source-text, target-text, $C_0$, and depth losses are weighted 0.4, 0.6, 1.0, and 3.0; training-only text-to-text, text-to-$C_0$, and speaker-identity objectives are weighted 0.3, 0.5, and 0.1. We use a polynomially decayed learning rate initialized at $10^{-4}$. At inference, beam search with a beam size of 5 generates the unified text--$C_0$ sequence, after which the depth decoder greedily predicts $C_1$--$C_{15}$ in parallel over time and sequentially across codebooks.\vspace{-2mm}

\section{Results \& Discussion}
\label{sec:results}
\noindent\textbf{Training dataset analysis.}
Table~\ref{tab:data_statistics} summarizes the training data.
Table~\ref{tab:pair_analysis} shows that \textsc{Raw-Dub}
already provides stronger content, prosody, and timing
correspondence than CVSS-T, while its main limitation is
speaker mismatch.
Applying VC substantially improves Vsim and further increases
A.PCP, with smaller gains in Rate and Pause.
Without manual transcripts, we compare pre- and post-VC
ASR outputs, obtaining approximately 8.3\% WER and 5.2\% CER
in both settings. 
The small changes in mean BLASER suggest largely preserved
cross-lingual semantic correspondence at the corpus level.
As a negative control, randomly shuffling the source--target
pairings in \textsc{VC-Dub} causes large drops in BLASER,
A.PCP, Rate, and Pause, indicating that the observed
correspondence is pair-specific rather than a corpus artifact.
Together, professional dubbing and VC yield our holistic parallel
supervision across content, prosody, timing, and speaker identity.


\begin{table}[t]

\centering
\scriptsize
\vspace{-2.5mm}
\setlength{\tabcolsep}{2.6pt}
\caption{Statistics of the speech-to-speech training data. Average duration is reported as source $|$ target.}
\renewcommand{\arraystretch}{0.58}
\resizebox{1.0\columnwidth}{!}{
\begin{tabular}{llccccc}
\toprule[0.7pt]
\textbf{Pair} & \textbf{Dataset} & \textbf{\#Pairs} 
& \textbf{Src. Hrs} & \textbf{Tgt. Hrs} & \textbf{Total Hrs} & \textbf{Avg. Dur. (s)} \\
\midrule
\multirow{2}{*}{En-De}
& CVSS-T  & 127,250 & 128.19 & 122.16 & 250.35 & 3.63 $|$ 3.46 \\
& \textsc{VC-Dub} & 131,399 & 127.40 & 135.82 & 263.22 & 3.49 $|$ 3.72 \\
\midrule
\multirow{2}{*}{En-Es}
& CVSS-T  & 78,963  & 79.76  & 72.77  & 152.53 & 3.64 $|$ 3.32 \\
& \textsc{VC-Dub} & 79,200 & 84.61 & 86.52 & 171.13 & 3.85 $|$ 3.93 \\
\bottomrule[0.7pt]
\end{tabular}
}
\vspace{-5mm}
\label{tab:data_statistics}
\end{table}
\begin{table}[t]
\vspace{-2mm}
\centering
\scriptsize
\setlength{\tabcolsep}{1.9pt}
\renewcommand{\arraystretch}{1.05}
\caption{Pair-level correspondence and speech-quality results for CVSS-T, \textsc{Raw-Dub}, and \textsc{VC-Dub}.
Shuffled \textsc{VC-Dub} pairs provide a negative control. DNSMOS$_{\mathrm{Pro}}$ is reported as source (En) $|$ target (De/Es).}
\resizebox{\columnwidth}{!}{%
\begin{tabular}{llcccccc}
\toprule
\textbf{Dir.}
& \textbf{Dataset}
& \textbf{BLASER}$\uparrow$
& \textbf{A.PCP}$\uparrow$
& \textbf{Rate}$\uparrow$
& \textbf{Pause}$\uparrow$
& \textbf{Vsim}$\uparrow$
& \textbf{DNSMOS$_{\mathrm{Pro}}$}$\uparrow$ \\
\midrule

\multirow{4}{*}{\shortstack{En\\$\downarrow$\\De}}
& CVSS-T
& 3.405
& 2.505
& 0.365
& 0.524
& 0.208
& 3.160 $|$ 2.797 \\

& \textsc{Raw-Dub}
& \textbf{3.634}
& 2.759
& 0.416
& 0.682
& 0.060
& 3.900 $|$ 3.684 \\

& \textsc{VC-Dub}
& 3.611
& \textbf{3.000}
& \textbf{0.435}
& \textbf{0.698}
& \textbf{0.406}
& \textbf{4.104 $|$ 3.684} \\

& w/ Shuffled
& 2.285
& 2.218
& -0.009
& 0.394
& 0.025
& 4.104 $|$ 3.684 \\

\midrule

\multirow{4}{*}{\shortstack{En\\$\downarrow$\\Es}}
& CVSS-T
& 3.400
& 2.537
& 0.372
& 0.550
& 0.194
& 3.173 $|$ 2.780 \\

& \textsc{Raw-Dub}
& \textbf{3.632}
& 2.920
& 0.417
& 0.676
& 0.046
& 3.913 $|$ 3.714 \\

& \textsc{VC-Dub}
& 3.606
& \textbf{3.183}
& \textbf{0.436}
& \textbf{0.686}
& \textbf{0.362}
& \textbf{4.091 $|$ 3.714} \\

& w/ Shuffled
& 2.199
& 2.345
& -0.003
& 0.361
& 0.027
& 4.091 $|$ 3.714 \\

\bottomrule
\end{tabular}
}

\label{tab:pair_analysis}
\label{tab:data_validation}
\vspace{-7mm}
\end{table}

\noindent\textbf{Effect of holistic parallel supervision on S2ST.}
As shown in Table~\ref{tab:main_s2st_results}, \textsc{VC-Dub}
improves expressive preservation across all three model families.
In particular, A.PCP and Vsim consistently improve for TransVIP,
Dub-S2ST, and DirectS2ST, with additional gains in timing and
naturalness depending on the architecture.
Dub-S2ST's speed adaptation mainly improves Rate, whereas
\textsc{VC-Dub} provides broader gains, supporting the benefit
of jointly matched supervision over matching a single temporal
attribute.

Absolute scores also reflect differences in model pretraining.
TransVIP retains stronger content performance, likely aided by
its SeamlessM4T-initialized speech-to-text translation backbone
\cite{le2024transvip}, whereas DirectS2ST trains its translation
and codec decoders from scratch to test whether paired speech
alone can support direct expressive supervision.
Despite this stricter setting, DirectS2ST+\textsc{VC-Dub}
achieves the best A.PCP and Pause scores in both directions,
with significant improvements over the strongest competing
systems.
Dub-S2ST obtains stronger absolute naturalness and speaker
similarity, plausibly benefiting from its unit-to-speech
synthesizer initialized from CosyVoice-300M
\cite{choi2025dubs2st}, which was trained with large-scale data; nevertheless, replacing CVSS-T with
\textsc{VC-Dub} still improves its A.PCP, Vsim, and translation
performance.
These results support the conclusion that holistic
parallel supervision benefits different expressive S2ST formulations. 
The larger expressive gains observed for DirectS2ST are consistent
with our direct-supervision hypothesis: because
the model learns the complete target speech representation
directly from paired speech, its expressive learning is especially
responsive to the quality of holistic source--target supervision.


\begin{table}[t]
\centering
\vspace{-2mm}
\scriptsize
\setlength{\tabcolsep}{0.55pt}
\caption{Expressive S2ST results on mExpresso in both translation directions. $^\dagger$ denotes significance between training sets within the same
model, and $^\ddagger$ against the strongest baseline in the same language
direction. Significance is determined by two-sided paired randomization
tests with 10,000 permutations and Holm--Bonferroni correction
($p_{\mathrm{adj}}<0.05$).
}
\renewcommand{\arraystretch}{0.96}
\resizebox{\columnwidth}{!}{%
\begin{tabular}{lllccccccc}
\toprule
\textbf{Dir.} & \textbf{Model} & \textbf{Train}
& \textbf{ASR-BLEU}$\uparrow$
& \textbf{BLASER}$\uparrow$
& \textbf{A.PCP}$\uparrow$
& \textbf{Rate}$\uparrow$
& \textbf{Pause}$\uparrow$
& \textbf{Vsim}$\uparrow$
& \textbf{NAT}$\uparrow$ \\
\midrule

\multirow{5}{*}{\shortstack{En\\$\downarrow$\\De}}
& TransVIP & CVSS-T
& 12.25 & 3.397
& 2.609 & 0.470 & 0.489
& 0.126 & 3.257 \\

& TransVIP & \textsc{VC-Dub}
& \textbf{13.84} & \textbf{3.399}
& 2.743$^{\dagger}$ & \textbf{0.600}$^{\dagger}$ & 0.549$^{\dagger}$
& 0.169$^{\dagger}$ & \textbf{3.472}$^{\dagger}$ \\
\cmidrule(lr){4-10}

& DirectS2ST & CVSS-T
& 6.01 & 3.234
& 2.162 & 0.334 & 0.687
& 0.114 & 3.339 \\

& DirectS2ST & \textsc{VC-Dub}
& 10.52$^{\dagger}$ & 3.373$^{\dagger}$
& \textbf{2.905}$^{\dagger,\ddagger}$ & 0.589$^{\dagger}$ & \textbf{0.752}$^{\dagger,\ddagger}$
& \textbf{0.217}$^{\dagger,\ddagger}$ & 3.450$^{\dagger}$ \\

\midrule

\multirow{9}{*}{\shortstack{En\\$\downarrow$\\Es}}
& TransVIP & CVSS-T
& 15.53 & 3.354
& 2.664 & 0.601 & 0.494
& 0.138 & 3.092 \\

& TransVIP & \textsc{VC-Dub}
& \textbf{18.96}$^{\dagger}$ & \textbf{3.433}$^{\dagger}$
& 2.803$^{\dagger}$ & \textbf{0.671}$^{\dagger}$ & 0.562$^{\dagger}$
& 0.158$^{\dagger}$ & 3.407$^{\dagger}$ \\

\cmidrule(lr){4-10}

& Dub-S2ST & CVSS-T
& 11.61 & 3.239
& 2.933 & 0.581 & 0.596
& 0.218 & 3.658 \\

& Dub-S2ST & CVSS-T$^{\mathrm{S}}$
& 11.89 & 3.226
& 2.938 & 0.608 & 0.572
& 0.214 & 3.631 \\

& Dub-S2ST & \textsc{VC-Dub}
& 15.21 & 3.321$^{\dagger}$
& 3.012$^{\dagger}$ & 0.595 & 0.612
& \textbf{0.238}$^{\dagger}$ & \textbf{3.759}$^{\dagger}$ \\

\cmidrule(lr){4-10}

& DirectS2ST & CVSS-T
& 10.00 & 3.153
& 2.238 & 0.433 & 0.652
& 0.105 & 3.361 \\

& DirectS2ST & \textsc{VC-Dub}
& 14.22$^{\dagger}$ & 3.259$^{\dagger}$
& \textbf{3.091}$^{\dagger,\ddagger}$ & 0.669$^{\dagger}$ & \textbf{0.733}$^{\dagger,\ddagger}$
& 0.205$^{\dagger}$ & 3.466$^{\dagger}$ \\

\bottomrule
\end{tabular}%
}
\label{tab:main_s2st_results}
\vspace{-6mm}
\end{table}

\begin{table}[!t]
\centering
\scriptsize
\setlength{\tabcolsep}{0.55pt}
\renewcommand{\arraystretch}{1.08}
\caption{
\textsc{VC-Dub} ablation of DirectS2ST on mExpresso.
}
\begin{tabular}{llccccccc}
\toprule
\textbf{Dir.}
& \textbf{Train}
& \textbf{ASR-BLEU}$\uparrow$
& \textbf{BLASER}$\uparrow$
& \textbf{A.PCP}$\uparrow$
& \textbf{Rate}$\uparrow$
& \textbf{Pause}$\uparrow$
& \textbf{Vsim}$\uparrow$
& \textbf{NAT}$\uparrow$ \\
\midrule

\multirow{2}{*}{\shortstack{En$\rightarrow$De}}
& \textsc{VC-Dub}
& 10.52 & 3.373
& 2.905$^{\dagger}$ & 0.589$^{\dagger}$ & 0.752
& 0.217$^{\dagger}$ & 3.450 \\

& \textsc{Raw-Dub}
& 9.37 & 3.339
& 2.712 & 0.455 & 0.741
& 0.065 & 3.430 \\

\midrule

\multirow{2}{*}{\shortstack{En$\rightarrow$Es}}
& \textsc{VC-Dub}
& 14.22 & 3.259
& 3.091$^{\dagger}$ & 0.669$^{\dagger}$ & 0.733
& 0.205$^{\dagger}$ & 3.466 \\

& \textsc{Raw-Dub}
& 14.53 & 3.266
& 2.824 & 0.555 & 0.730
& 0.058 & 3.470 \\

\bottomrule
\end{tabular}

\label{tab:ablation}
\end{table}

\begin{table}[!t]
\centering
\vspace{-5mm}
\caption{Expressiveness MOS with 95\% confidence intervals and
inter-rater reliability (ICC). Significance is assessed using
crossed linear mixed-effects models with system configurations as
a fixed effect and listener and utterance as random effects,
followed by Holm--Bonferroni correction.
$^\dagger p_{\mathrm{adj}}<0.05$.}

\label{tab:mos}
\scriptsize

\resizebox{\columnwidth}{!}{
\begin{tabular}{llcccc}
\toprule
Model & Training Data & MOS $\uparrow$ & 95\% CI
& ICC(A,1) & ICC(A,9) \\
\midrule

TransVIP
& CVSS-T
& 1.82
& [1.62, 2.01]
& \multirow{8}{*}{0.445}
& \multirow{8}{*}{0.878} \\

TransVIP
& \textsc{VC-Dub}
& 2.24$^\dagger$
& [2.03, 2.44]
& & \\

\cmidrule(lr){3-4}

Dub-S2ST
& CVSS-T
& 2.88
& [2.63, 3.12]
& & \\

Dub-S2ST
& \textsc{VC-Dub}
& \textbf{3.59}$^\dagger$
& \textbf{[3.34, 3.84]}
& & \\

\cmidrule(lr){3-4}

DirectS2ST
& CVSS-T
& 2.13
& [1.93, 2.33]
& & \\

DirectS2ST
& \textsc{VC-Dub}
& 3.42$^\dagger$
& [3.20, 3.64]
& & \\

\bottomrule
\end{tabular}
}
\vspace{-7mm}
\end{table}


\noindent\textbf{Contribution of voice conversion.}
Table~\ref{tab:ablation} compares DirectS2ST trained on the same aligned dubbing pairs with and without VC. \textsc{Raw-Dub} already provides effective expressive supervision but, as expected, does not preserve speaker identity. VC substantially improves speaker similarity and yields additional gains in A.PCP and Rate, while content, Pause, and naturalness remain comparable. These results indicate that VC primarily reduces speaker mismatch while preserving the other properties of the dubbed speech.


\noindent\textbf{Human evaluation.}
Table~\ref{tab:mos} reports human ratings of perceived
expressiveness. \textsc{VC-Dub} significantly improves MOS
across all three model families with high inter-rater reliability, confirming that the benefits of holistic
parallel supervision are also reflected in perceived
expressiveness. Among the evaluated systems, DirectS2ST shows
the largest MOS gain (+1.29), mirroring the automatic-metric
results and further suggesting that direct source-to-target
modeling can exploit holistic paired supervision particularly
effectively.
\vspace{-0.5mm}
\section{Conclusion}

We investigated holistic parallel supervision for expressive S2ST by combining professional dubbing, which provides human-performed correspondence in content, prosody, and timing, with multilingual voice conversion to reduce speaker mismatch while preserving natural target speech. Experiments on En--De and En--Es show improved expressive preservation across multiple S2ST systems, supported by human evaluation. The Raw-Dub ablation isolates the contribution of speaker matching, while DirectS2ST achieves the largest relative gains in content, prosody, timing, and perceived expressiveness, suggesting that direct target-speech prediction particularly benefits from holistic parallel supervision.
\clearpage
\begingroup
\small  
\bibliographystyle{IEEEbib}
\bibliography{strings,refs}
\end{document}